\documentclass[10pt,twocolumn,letterpaper]{article}

\usepackage[pagenumbers]{cvpr} % arXiv 模式已开启 (删除了前面的 %)
\usepackage{booktabs}
\usepackage{multirow}
\usepackage{makecell} 
\definecolor{cvprblue}{rgb}{0.21,0.49,0.74}
\usepackage[pagebackref,breaklinks,colorlinks,allcolors=cvprblue]{hyperref}

\def\paperID{} % *** Enter the Paper ID here
\def\confName{CVPR}
\def\confYear{2026}
\usepackage[table]{xcolor} % 导言区添加
\title{SegMoTE: Token-Level Mixture of Experts for Medical Image Segmentation}

\author{
	Yujie Lu$^{1*}$, \quad
	Jingwen Li$^{2*}$, \quad
	Sibo Ju$^{3}$, \quad
	Yanzhou Su$^{4}$ \\
	He Yao$^{1}$, \quad
	Yisong Liu$^{1}$, \quad
	Min Zhu$^{1\dagger}$, \quad
	Junlong Cheng$^{1\dagger}$ \\[2mm]
	$^1$Sichuan University \quad
	$^2$Xinjiang University \\
	$^3$Fuzhou University \quad
	$^4$Alibaba DAMO Academy \\[1mm]
	{\small $^*$Equal contribution \quad $^\dagger$Corresponding authors} \\[1mm]
	{\small \url{https://github.com/InMyDreammer/SegMoTE}}
}

\begin{document}
	\maketitle
	\begin{abstract}
Medical image segmentation is vital for clinical diagnosis and quantitative analysis, yet remains challenging due to the heterogeneity of imaging modalities and the high cost of pixel-level annotations. Although general interactive segmentation models like SAM have achieved remarkable progress, their transfer to medical imaging still faces two key bottlenecks: (i) the lack of adaptive mechanisms for modality- and anatomy-specific tasks, which limits generalization in out-of-distribution medical scenarios; and (ii) current medical adaptation methods fine-tune on large, heterogeneous datasets without selection, leading to noisy supervision, higher cost, and negative transfer. To address these issues, we propose SegMoTE, an efficient and adaptive framework for medical image segmentation. SegMoTE preserves SAM’s original prompt interface, efficient inference, and zero-shot generalization while introducing only a small number of learnable parameters to dynamically adapt across modalities and tasks. In addition, we design a progressive prompt tokenization mechanism that enables fully automatic segmentation, significantly reducing annotation dependence. Trained on MedSeg-HQ, a curated dataset less than 1\% of existing large-scale datasets, SegMoTE achieves SOTA performance across diverse imaging modalities and anatomical tasks. It represents the first efficient, robust, and scalable adaptation of general segmentation models to the medical domain under extremely low annotation cost, advancing the practical deployment of foundation vision models in clinical applications.
\end{abstract}

	\section{Introduction}
\label{sec:intro}

Medical image segmentation, as an important application of computer vision in the medical field, faces the core challenges of accurate segmentation and generalization across different imaging modalities such as CT, MRI, and X-ray \cite{rui2025multi,wang2025toward,huo2025generative,ma2025steady}. Due to the heavy reliance on medical professionals for pixel-level lesion annotation and the privacy and compliance restrictions on data, acquiring high-quality annotated data is extremely costly. In this context, interactive medical image segmentation methods, by incorporating user interaction, effectively alleviate the annotation burden and improve data iteration efficiency. However, existing interactive methods still have limited adaptability in cross-modal scenarios and struggle to achieve stable generalization across different imaging modalities. Therefore, how to efficiently leverage natural image pre-trained models and enable their effective transfer and adaptation across diverse medical imaging domains is a key issue that remains to be addressed.

\begin{figure}[t]
  \centering
  % 调整图片宽度为 50% 页面宽度
  \includegraphics[width=0.48\textwidth]{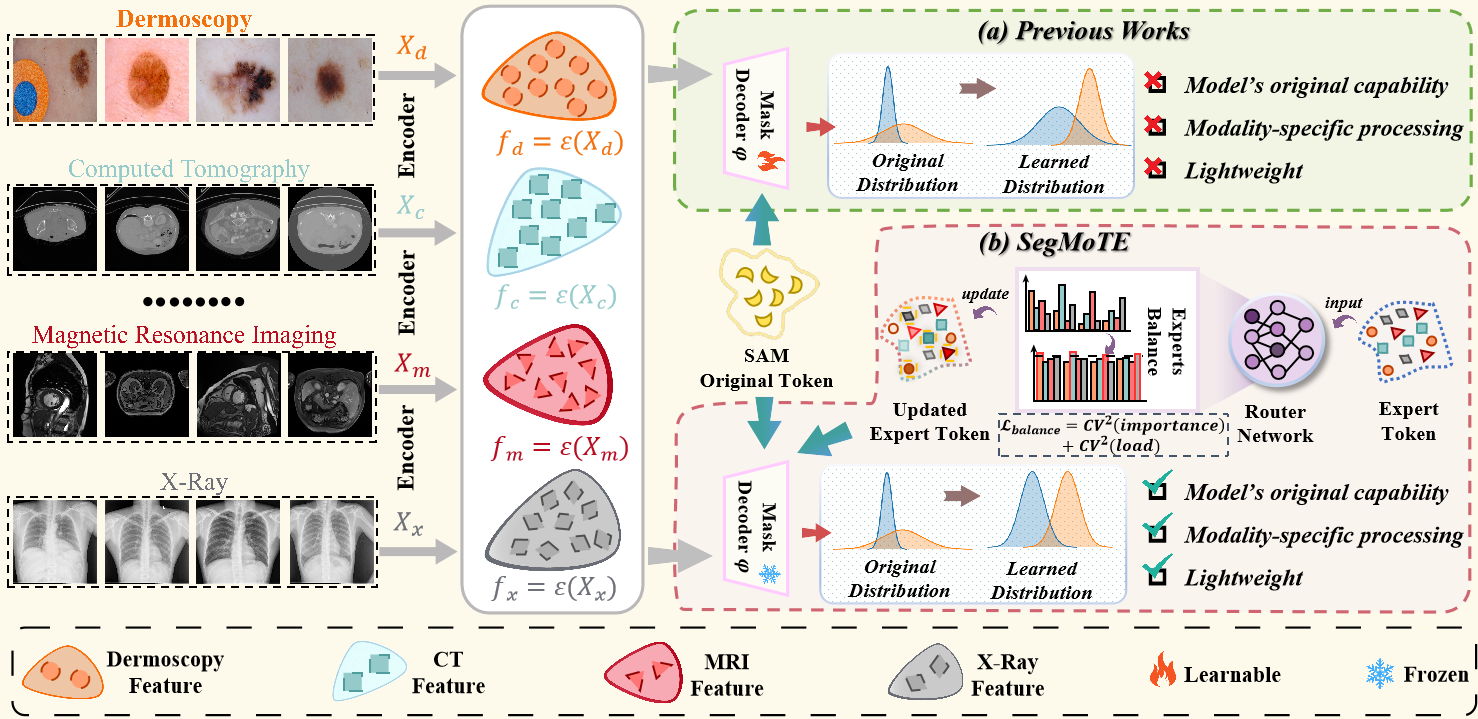}
  \caption{\textbf{SegMoTE vs. Previous Works.} The heterogeneous data $X$ is first processed by the encoder $\varepsilon$ to extract the feature representation $f$. (a) Previous methods typically perform full fine-tuning of the mask decoder or parameter-efficient fine-tuning, leading to distribution shift from the pretrained model. (b) SegMoTE introduces a token-level mixture of experts mechanism that dynamically selects modality-adaptive expert tokens while keeping the mask decoder frozen. The process is guided by the load balancing loss $L_{balance}$, is constrained using the squared coefficient of variation $({CV}^2)$. This design preserves SAM’s original capability, enhances modality-specific adaptability, and maintains a lightweight architecture.}
  \label{fig:first}
\end{figure}

As the artificial intelligence community evolves, a series of powerful foundational models such as CLIP \cite{radford2021learning}, SAM \cite{kirillov2023segment,ravisam}, and DINO \cite{zhangdino,oquab2023dinov2,simeoni2025dinov3} continue to emerge. Through large-scale pre-training and cross-modal learning, these models have significantly advanced interactive medical image segmentation technology. Notably, the Segment Anything Model (SAM) demonstrates exceptional performance across diverse scenarios due to its universal segmentation capabilities. Subsequently, researchers have fine-tuned the SAM series models to accommodate the diversity of medical images. For example, MedSAM \cite{ma2024segment} introduces a foundation model enabling universal and robust medical image segmentation across modalities, while IMIS \cite{cheng2025interactive} constructed the large-scale IMed-361M dataset and trained SAM on it, obtaining favorable segmentation results. 

Despite recent progress, two challenges persist: (i) the absence of modality- and task-aware adaptation, which limits out-of-distribution generalization; and (ii) undifferentiated data aggregation introduces excessive supervision noise and inherent redundancy, which in turn hinders learning efficiency and impedes the acquisition of effective knowledge. As shown in Fig.\ref{fig:first} (a), previous works generally feed heterogeneous multimodal data directly into SAM and adapt it to downstream tasks through full-parameter fine-tuning \cite{ma2024segment}, decoder layer fine-tuning \cite{cheng2025interactive}, or parameter-efficient fine-tuning \cite{zhongconvolution}. However, with the continuous increase in data scale and modality diversity, the original output tokens of SAM become progressively homogenized during training, leading to insufficient discriminability among modality specific representations and weakening the model’s ability to capture cross-modal semantic differences. Moreover, chasing performance by expanding datasets induces distribution shift: representations drift toward the new data distribution and compromise the model’s original capabilities. In effect, progress has devolved into a race for data scale rather than advances in representation design.

To address these challenges, we propose SegMoTE (Segmentation with Mixture of Token Experts), an interactive medical image segmentation framework built upon the Mixture of Experts (MoE) paradigm \cite{jacobs1991adaptive,shazeer2017outrageously,lepikhingshard}. As illustrated in Fig.\ref{fig:first} (b), SegMoTE introduces token-level expert selection, dynamically activating the most suitable expert tokens for each imaging modality to achieve modality specific representation and adaptive processing. This design enables independent feature extraction across modalities, effectively mitigating the limitations of conventional methods in handling modality heterogeneity. We freeze the SAM encoder and train only lightweight MoTE, preserving SAM’s original capability while extending it to new datasets. This achieves a unified balance of efficiency, stability, and adaptability. To support training, we construct MedSeg-HQ, a curated multimodal dataset containing about 0.15M high-quality masks. Despite its relatively small scale, MedSeg-HQ provides sufficient diversity and representativeness, enabling SegMoTE to achieve robust, high performance segmentation even under limited data conditions. 

Additionally, in sparse class segmentation tasks ,such as the ISIC dermatology segmentation dataset \cite{gutman2016skin,codella2018skin}, samples typically contain only background and a single target class. Existing visual prompting strategies rely on user provided guidance, increasing the operational burden. To reduce this dependency, we propose Progressive Prompt Tokenization (PPT). PPT alternates between mask and text prompts to progressively guide prompt tokens toward foreground and background regions. This process gradually transforms latent features into semantically aligned token representations, enabling accurate segmentation without any human intervention during inference.

To validate the effectiveness of our approach, we conducted experiments across multiple datasets encompassing diverse modalities, with results consistently demonstrating its superiority. In summary, our main contributions are as follows:

\begin{itemize}
    \item We propose the SegMoTE framework, which preserves SAM’s strong zero-shot capability and flexibility while achieving modality adaptive and precise medical image segmentation through dynamic selection and updating of expert tokens, using only minimal additional parameters and training data.
    \item We introduce MedSeg-HQ, a multimodal medical segmentation dataset. Despite being only 1\% the size of existing datasets, it boosts model generalization and transferability with minimal supervision, setting a new benchmark for interactive medical image segmentation.
    \item We propose Progressive Prompt Tokenization, which uses randomly sampled mask and text prompts to guide prompt tokens progressively toward foreground and background regions, enabling efficient, interaction-free few-class segmentation.
    \item Extensive experiments on both in-domain and out-of-domain datasets demonstrate that our method improves 1\% to 6\% over the second best approach by training only 17M parameters on 0.15M dataset.
\end{itemize}

%-------------------------------------------------------------------------

	\section{Related Work}

%-------------------------------------------------------------------------
\subsection{Medical Image Segmentation Based on SAM}
SAM is the first universal image segmentation foundation model, demonstrating strong zero-shot generalization capabilities in natural image segmentation \cite{zhang2023comprehensive,ke2023segment,shu2025tinysam,xiong2024efficientsam}. However, due to significant differences between natural and medical images in terms of modal diversity and anatomical complexity, directly applying pre-trained SAM to medical image segmentation poses substantial challenges. The COSMOS 1050K dataset introduced by Huang et al. \cite{huang2024segment} validated the limitations of SAM in medical images, particularly its instability when handling complex anatomical structures and irregular boundaries. MedSAM \cite{ma2024segment} fine-tunes SAM on medical datasets, demonstrating its potential for medical image segmentation, though it still faces challenges related to sparse annotations. 

SAM-Med2D \cite{cheng2023sammed2d} introduced a comprehensive solution by collecting 4.6M images and 19.7M masks to construct a large-scale medical image dataset. By fine-tuning SAM with diverse prompts (boxes, points, masks), it substantially improved performance in multimodal and multi-organ segmentation. Furthermore, IMed-361M \cite{cheng2025interactive} introduced a large-scale, densely annotated interactive medical image segmentation benchmark and developed IMIS-Net, setting new SOTA records across multiple external datasets. This demonstrated the critical role of dense annotations for SAM in achieving “segment everything” capabilities in medical images. Despite significant progress in dataset construction and model design, existing methods often rely on expanding the dataset size to enhance performance, overlooking issues such as data redundancy and modality discrepancies, which may lead to a degradation in representational capability.

%-------------------------------------------------------------------------
\subsection{Mixture of Experts}
The Mixture of Experts (MoE) model has gained significant attention in deep learning for its ability to enhance model performance while maintaining computational efficiency by integrating multiple expert networks with a gating mechanism \cite{sun2025controllable,zhang2025mone,puigcerversparse}. This framework has shown success in large-scale language modeling, such as DeepSeek \cite{dai2024deepseekmoe,guo2025deepseek,liu2024deepseek}, Hunyuan-LaRGE \cite{sun2024hunyuan}, and LLaMA-MoE \cite{zhu2024llama,qu2024llama}. MoE has also been extended to computer vision tasks, including image classification \cite{zhao2025mexd}, object detection \cite{wang2025object}, and semantic segmentation \cite{pavlitska2025extracting}, demonstrating its broad applicability.

In medical image analysis, MoE has been adopted for its ability to handle multimodal and multi-task learning. Examples include MoSE, which improves segmentation accuracy with shape prior experts \cite{wei2025mixture}. M4oE, which enhances generalization with dynamic gating in Swin-UNet \cite{jiang2024m4oe}, and PAMoE, which integrates multi-scale feature fusion in histopathology images \cite{wu2025learning}. ConvLoRA introduces visual inductive biases to model dynamic feature selection mechanisms \cite{zhongconvolution}. However, most existing methods rely on a unified output representation, failing to address the differences between imaging modalities and tasks. In contrast, our proposed architecture uses modality-aware routing to effectively allocate inputs to dedicated expert paths, enabling more efficient and discriminative representation learning for multimodal medical images.

%-------------------------------------------------------------------------

%-------------------------------------------------------------------------

%-------------------------------------------------------------------------

	\section{Method}

\begin{figure*}[t]
  \centering
  \includegraphics[width=1\linewidth]{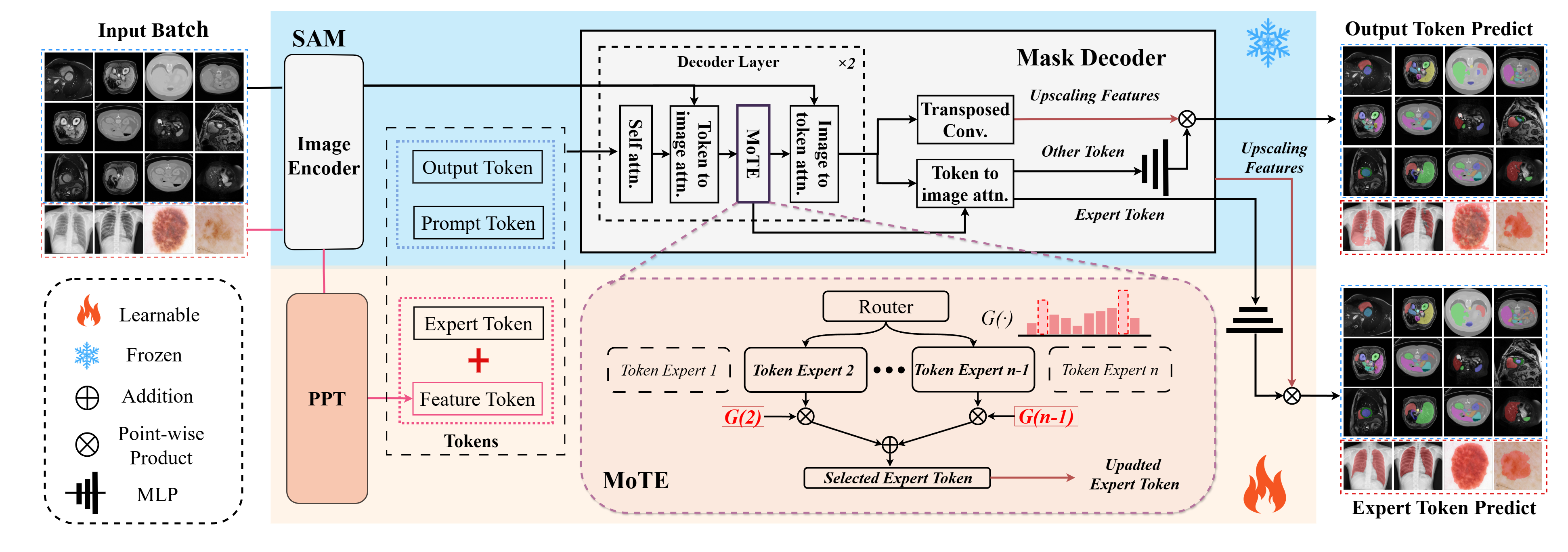}
  \caption{\textbf{Overview of the proposed SegMoTE framework.}
  	SegMoTE extends SAM by introducing a token-level expert routing mechanism to enable adaptive multimodal medical image segmentation. The frozen SAM encoder extracts modality-agnostic embedding representations, while the progressive prompt tokenization transforms latent features into semantically aligned feature tokens. These tokens interact with the decoder layers and MoTE for dynamic expert selection and adaptive token updates.
  }
  \label{fig:overview}
\end{figure*}

\subsection{Overview}
As illustrated in Fig.\ref{fig:overview}, the proposed SegMoTE framework extends SAM with token-level expert routing for adaptive multimodal medical image segmentation. The frozen SAM encoder first extracts image embeddings, which serve as modality-agnostic representations for subsequent processing. The PPT, applied only to the few-class datasets highlighted by the red dashed box such as ISIC \cite{codella2019skin,codella2018skin} and SZ-CXR \cite{stirenko2018chest}, converts latent feature maps into semantically aligned feature token. These feature tokens are concatenated with other tokens and fed into decoder layer 1 and decoder layer 2 of the mask decoder. Each decoder layer performs self-attention, followed by token to image attention to interact with image embeddings, and then passes the results to the MoTE for dynamic expert selection and token updating. The updated expert tokens are used for final prediction, enabling modality-specific processing within a unified and lightweight framework while preserving SAM’s original capability.

\subsection{Expert Token}
In the original SAM mask decoder design shown in Fig.\ref{fig:overview}, mask prediction is based on output tokens, which are processed through dynamically predicted MLP weights and then interact with the mask features via pointwise multiplication. However, this process shows limited adaptability when handling heterogeneous medical image modalities. To address this, we introduce the expert token in our interactive medical segmentation framework, SegMoTE. By assigning dedicated tokens for different modalities, we enhance modality adaptability while retaining SAM’s original unified output modeling capability.

Specifically, we introduce a set of learnable expert tokens with a dimension of [N$\times$256], where N depends on the number of input modalities or task complexity. These tokens are concatenated with the original SAM output tokens [4$\times$256] and prompt tokens along the sequence dimension and fed into the mask decoder. The expert tokens first interact with other tokens via self attention in each decoder layer, then undergo feature updates through bidirectional attention (token to image and image to token). In the token to image phase, expert tokens integrate visual features from the image modality, geometric and semantic information from the prompt tokens, and mask representations from other tokens. They are then sent to the MoTE for dynamic weight updating, detailed in \cref{sec:MoTE}, before interacting in the image to token attention layer. Finally, all tokens are processed through a token to image attention layer and MLP layer, element-wise multiplied with the image features to generate the segmentation mask prediction. Throughout this process, only the selected expert token is utilized for prediction, enabling differentiated processing of multimodal images within the same batch.

\subsection{Mixture of Token Experts}
\label{sec:MoTE}
Although expert tokens provide differentiated representations for multimodal inputs, efficiently utilizing these representations and selecting the most suitable token for each image during inference is crucial for enhancing the model's adaptability. To address this, we introduce the MoTE (Mixture of Token Experts) mechanism that achieves dynamic expert selection and fusion at the token level, maintaining modality distinction while balancing feature diversity.

The core of MoTE lies in providing adaptive token selection capabilities for each image. As shown in Fig.\ref{fig:overview}, given an input expert token $\mathbf{x}\in\mathbb{R}^{B\times T\times D}$, where $B$ is the batch size, $T$ is the number of tokens per image, and $D$ is the token dimension. We first compute the logits for each token across all experts using the router:
\begin{equation}
  \mathrm{L=XW_g}\in\mathbb{R}^{B\times T\times E},
  \label{eq:important}
\end{equation}
where $\mathrm{W}_{\mathrm{g}}$ is the gating weight matrix, $E$ is the number of experts. During training, we adopt the noisy top-k gating approach, injecting noise into the logits to encourage exploration and prevent expert routing from prematurely converging to a single mode:
\begin{equation}
  \tilde{\mathbf{L}}=\mathbf{L}+(\mathrm{softplus}(\mathbf{XW}_n)+\varepsilon)\odot\mathbf{Z},\quad\mathbf{Z}\sim\mathcal{N}(0,1),
  \label{eq:important}
\end{equation}
where $W_n$ is the parameter matrix predicted by the noise scale, and $\varepsilon$ is set to 1e-2 for numerical stability to prevent zero standard deviations. $Z$ is a gaussian random noise tensor with the same shape as the logits. Subsequently, for the t-th token in the b-th image of the batch, we denote its top-k expert score as:
\begin{equation}
  \mathbf{s}_{b,t}= \begin{pmatrix} \tilde{L}_{b,t,i_1},\tilde{L}_{b,t,i_2},...,\tilde{L}_{b,t,i_k} \end{pmatrix}\in\mathbb{R}^k,\quad i_j\in\mathcal{S}_{b,t},
  \label{eq:important}
\end{equation}
where ${s}_{b,t}$ is the top-k expert index set for the token, we use the maximum score as the confidence level for that token. We then compare the confidence levels of different tokens to determine the final selected token, using its corresponding expert as the final routing target. Before determining the optimal expert index $\mathrm{idx}_{b,t}$, we further utilize the maximum logit to obtain the token confidence score $\mathrm{c}_{b,t}$:
\begin{equation}
  c_{b,t}=\max_{j=1,...,k}\mathbf{s}_{b,t}[j],\quad\mathrm{idx}_{b,t}=\arg\max_{j=1,...,k}\mathbf{s}_{b,t}[j],
  \label{eq:important}
\end{equation}
$c_{b,t}$ represents the token's highest score among candidate experts, where ${idx}_{b,t}$ denotes its index, thereby obtaining the token weight:
\begin{equation}
  G(\cdot)_{b,t}=\mathrm{softmax}(c_{b,t}),\quad a_{b,t}\in(0,1],
  \label{eq:important}
\end{equation}

We use $G(\cdot)_{b,t}$ as a reliability metric to explicitly weight token representations, amplifying high-confidence tokens while suppressing low-confidence tokens:
\begin{equation}
  \tilde{\mathbf{z}}_{b,t}=G(\cdot)_{b,t}\cdot\mathbf{h}_{b,t}^{(\mathrm{idx}_{b,t})},
  \label{eq:important}
\end{equation}
here, $\mathbf{h}_{b,t}^{(\mathrm{idx}_{b,t})}$ denotes the representation of $token_{(b, t)}$ obtained through the optimal expert branch and subsequent mapping. In Fig.\ref{fig:overview}, we present this using a top-2 approach, with token expert 2 and token expert n-1 serving as the activated experts. The updated expert token and the selected token index are fed back into the mask decoder. To prevent overcrowding or prolonged inactivity among experts, we introduce a load balancing loss during the routing phase. Specifically, the importance and load of expert $e$ are defined as:
\begin{equation}
  \mathrm{imp}_e=\sum_{b,t}\mathbf{G}_{b,t,e},\quad\mathrm{load}_e=\sum_{b,t}\mathbf{1}\left(\mathbf{G}_{b,t,e}>0\right),
  \label{eq:important}
\end{equation}
the loss $L_{balance}$ based on the coefficient of variation is calculated as follows:
\begin{equation}
	\mathcal{L}_{\mathrm{balance}}=\mathrm{CV}^2\left(\{\mathrm{imp}_e\}_{e=1}^E\right)+\mathrm{CV}^2\left(\{\mathrm{load}_e\}_{e=1}^E\right),
	\label{eq:important}
\end{equation}

\begin{equation}CV^2=\frac{\operatorname{std}(x)^2}{\left(\frac{1}{N}\sum_{i=1}^Nx_i\right)^2}\end{equation}
this constraint encourages more balanced utilization of experts, thereby enhancing overall training stability and generalization capabilities.

Ultimately, the model makes predictions using only the reinforced tokens routed through the network. This process, driven by the deterministic routing determined by $\mathrm{id}x_{b,t}$ and the confidence weighted routing determined by $\mathrm{G}(\cdot)_{b,t}$, enables expert token selection and information focusing.

\begin{figure}[t]
	\centering
	% 调整图片宽度为 50% 页面宽度
	\includegraphics[width=0.47\textwidth]{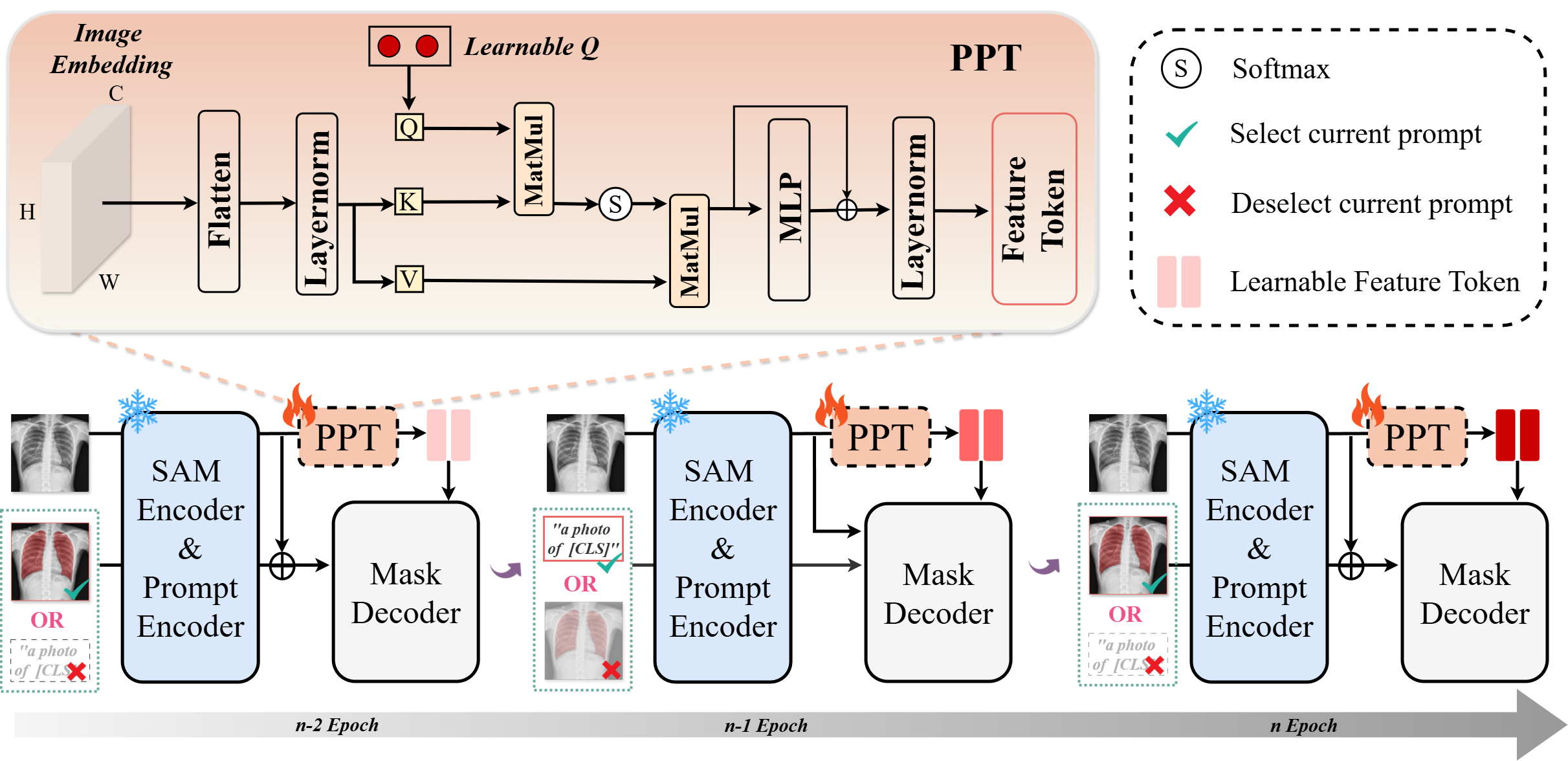}
	\caption{\textbf{Architecture of the Progressive Prompt Tokenization.} 
		By randomly selecting mask and text prompts as foreground priors, the learnable query $Q$ captures the relationship between the foreground and background by performing attention on the normalized image features.}
	\label{fig:PPT}
\end{figure}

\subsection{Progressive Prompt Tokenization}
To reduce reliance on manual annotations and enable adaptive prompting for medical image segmentation, we propose the progressive prompt tokenization mechanism. PPT treats mask and text prompts as concrete representations of foreground information. As shown in Fig.\ref{fig:PPT}, unlike traditional prompt-based segmentation methods that rely on user specified inputs, PPT randomly samples mask and text prompts to guide learnable query $Q$ via multi-head attention toward normalized image features. During training, the feature token gradually learns to distinguish foreground from background, capturing key distributional cues. The attention enhanced representations are further refined through MLP projection and residual fusion to generate feature-conditioned prompt tokens. These dynamic tokens serve as adaptive prompts that remain contextually aligned with imaging modalities and anatomical structures, enabling fully automatic segmentation while preserving SAM’s generalization ability and efficiency.

This method performs well on binary classification datasets with clear foreground-background distinction, such as ISIC and SZ-CXR, reducing reliance on explicit visual prompts. However, in multi-class segmentation tasks (e.g., multi-organ segmentation), class interference complicates accurate prompt token mapping. Therefore, this paper focuses on applying the PPT method to binary classification tasks to maximize its simplicity and inference efficiency.

\subsection{Loss Function}
After obtaining the prediction result $y^{E}$ through the expert token, we need to calculate the segmentation loss of the model. For segmentation tasks, the loss function is computed using dice loss\cite{milletari2016v}, which measures the discrepancy between the model's prediction and the true label $y$:
\begin{equation}
  L_{\mathrm{seg}}(y^E,\mathbf{y})=1-\frac{2\sum_iy_i^Ey_i}{\sum_iy_i^E+\sum_iy_i},
  \label{eq:important}
\end{equation}
to comprehensively account for partitioning loss and load balancing loss, we must also incorporate the load balancing loss $\mathcal{L}_{\mathrm{balance}}$ into the final total loss function:

\begin{equation}
  \mathcal{L}_{\mathrm{total}}=\mathcal{L}_{\mathrm{seg}}+\lambda_{\mathrm{balance}}\cdot\mathcal{L}_{\mathrm{balance}},
  \label{eq:important}
\end{equation}
$\lambda_{\mathrm{balance}}$ is a hyperparameter that controls the influence of load balancing loss on the total loss. To ensure that the load balancing loss does not overwhelm the segmentation task loss, we set it to a small value 0.01 during training. This approach guarantees expert load balancing while ensuring the model continues to prioritize optimizing segmentation task performance.

\begin{figure*}[t]
  \centering
  \includegraphics[width=0.9\linewidth]{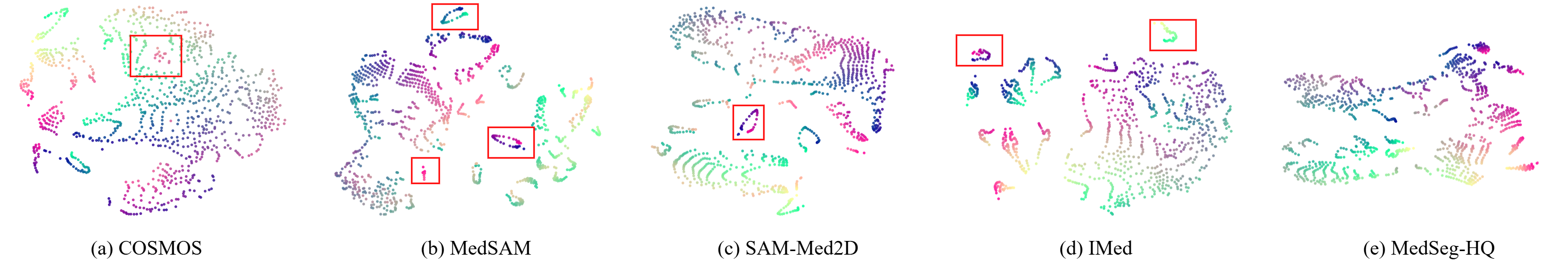}
  \caption{\textbf{Feature distribution comparison across datasets.} Feature embeddings extracted from the frozen SAM encoder are visualized after dimensionality reduction. The red boxes highlight regions where features exhibit overlap and unsmooth transitions. Compared with other datasets, MedSeg-HQ presents smoother and more continuous feature distributions, indicating higher consistency. % 这里写图注
  }
  \label{fig:tsne}
\end{figure*}

\subsection{MedSeg-HQ Data Construction}
To efficiently train the SegMoTE, we created a new dataset, MedSeg-HQ, consisting of 154,569 high-quality annotations, instead of using existing large-scale medical image segmentation datasets like COSMOS or IMed-361M. MedSeg-HQ integrates 12 public datasets, including CHAOS (T1), CHAOS (T2) \cite{kavur2021chaos,kavur2019chaos}, ISIC2016–2018 \cite{codella2018skin,gutman2016skin,codella2019skin}, BTCV \cite{landman2015miccai}, ACDC \cite{bernard2018deep}, AMOS (CT), AMOS (MRI) \cite{ji2022amos}, WORD \cite{luo2022word}, Totalsegmentator (CT) \cite{wasserthal2023totalsegmentator}, and SZ-CXR \cite{stirenko2018chest}, covering six modalities and over 100 semantic categories. To ensure high data quality, we developed a quality assessment system with five experts. We evaluated images based on clarity, contrast, entropy, foreground ratio, and connected regions, prioritizing those with higher scores. All datasets underwent random sampling and expert cross-validation, resulting in a MedSeg-HQ dataset with both diversity and quality. More details are in the supplementary materials. To compare with other datasets, we froze the SAM encoder and visualized the features using dimensionality reduction. As shown in Fig.\ref{fig:tsne}, MedSeg-HQ features show smooth transitions, while other datasets display more scatter and abrupt changes.

\begin{figure*}[t]
  \centering
  \includegraphics[width=0.8\linewidth]{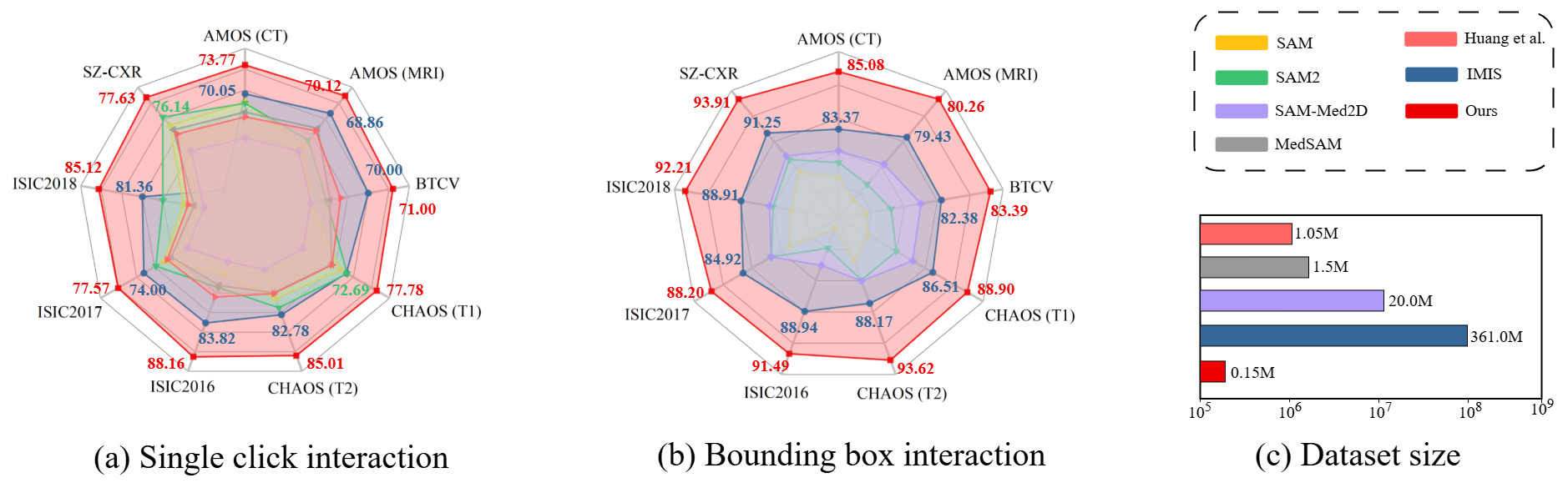}
  \caption{\textbf{In-domain segmentation results across datasets.} (a) and (b) show the Dice coefficient comparisons under single click and bounding box interactions, respectively. (c) illustrates the training dataset sizes used by different methods, where SegMoTE achieves performance improvement by optimizing the annotation quality of the data.
  }
  \label{fig:indomain}
\end{figure*}

\begin{table*}[t]
	\centering
	\caption{\textbf{Performance comparison on three out-of-domain datasets.}
		Experimental results of SegMoTE in comparison to other methods under bounding-box interactions. Best results are highlighted in \textbf{\color[HTML]{FE0000}{red}}, and second best results are highlighted in \textbf{\color[HTML]{0073e6}{blue}}.
		}
	\label{tab:out}
	\setlength{\tabcolsep}{5pt} % 列间距（按需调整/删除）
	\begin{tabular}{@{}llccccccc@{}}
		\toprule
		Dataset & Category & SAM & SAM 2 & MedSAM & Huang et al. & SAM-Med2D & IMIS&Ours \\
		\midrule
		\multirow{1}{*}{ISLES}
		& Ischemic Stroke Lesion & 55.00 & 59.17 & 58.02 & 56.54 & 67.93 & \textbf{\color[HTML]{0073e6}{71.24}}&\textbf{\color[HTML]{FE0000}{77.30}} \\
		\midrule
		\multirow{5}{*}{SegThor}
		& Esophagus & 68.94 & 77.69 & 38.59 & 68.47 & 73.92 & \textbf{\color[HTML]{0073e6}{80.82}}&\textbf{\color[HTML]{FE0000}{81.84}} \\
		& Heart     & 80.28 & \textbf{\color[HTML]{FE0000}{81.73}} & 68.09 & 78.21 & \textbf{\color[HTML]{0073e6}{80.34}} & 72.31&74.51 \\
		& Aorta     & 74.52 & 76.16 & 53.73 & 86.25 & 83.76 & \textbf{\color[HTML]{FE0000}{88.67}}&\textbf{\color[HTML]{0073e6}{87.40}} \\
		& Trachea   & \textbf{\color[HTML]{0073e6}{82.48}} & \textbf{\color[HTML]{FE0000}{83.64}} & 55.97 & 81.78 & 78.18 & 80.28&80.53 \\
		\cmidrule(lr){2-9}
		& \textbf{Average} & 76.55 & 79.81 & 54.09 & 78.67 & 79.06 & \textbf{\color[HTML]{0073e6}{80.52}}&\textbf{\color[HTML]{FE0000}{83.39}} \\
		\midrule
		\multirow{13}{*}{\shortstack[l]{Totalsegmentator\\(MRI)}}
		& Adrenal gland        & 63.90 & 65.76 & 48.67 & 55.53 & 54.59 & \textbf{\color[HTML]{0073e6}{67.00}}&\textbf{\color[HTML]{FE0000}{67.05}} \\
		& Aorta                & 70.47 & 72.61 & 50.01 & 74.73 & 74.61 & \textbf{\color[HTML]{0073e6}{75.11}} &\textbf{\color[HTML]{FE0000}{78.66}}\\
		& Colon                & 65.59 & \textbf{\color[HTML]{0073e6}{67.40}} & 38.18 & 49.85 & 50.98 & 65.14 &\textbf{\color[HTML]{FE0000}{68.23}}\\
		& Duodenum             & 64.08 & 65.34 & 52.73 & 62.08 & 62.73 & \textbf{\color[HTML]{0073e6}{66.05}} &\textbf{\color[HTML]{FE0000}{66.57}}\\
		& Gallbladder          & 70.12 & 71.89 & 65.00 & \textbf{\color[HTML]{0073e6}{75.80}} & \textbf{\color[HTML]{FE0000}{76.11}} & 71.17 &72.21\\
		& Iliopsoas            & 60.53 & 62.72 & 59.46 & 61.97 & 62.68 & \textbf{\color[HTML]{FE0000}{65.26}} &\textbf{\color[HTML]{0073e6}{64.31}}\\
		& Kidney               & 67.73 & 69.64 & 51.24 & 66.44 & 66.05 & \textbf{\color[HTML]{0073e6}{72.54}} &\textbf{\color[HTML]{FE0000}{74.80}}\\
		& Inferior vena cava   & 72.60 & 75.96 & 55.81 & 71.83 & 71.42 & \textbf{\color[HTML]{0073e6}{77.26}} &\textbf{\color[HTML]{FE0000}{78.21}}\\
		& Liver                & 77.16 & 78.73 & 65.41 & 76.59 & 78.27 & \textbf{\color[HTML]{0073e6}{80.32}} &\textbf{\color[HTML]{FE0000}{80.42}}\\
		& Pancreas             & 54.10 & \textbf{\color[HTML]{0073e6}{58.57}} & 39.73 & 56.70 & 57.19 & \textbf{\color[HTML]{FE0000}{60.21}} &58.63\\
		& Spleen               & 71.45 & 73.08 & 57.39 & 74.85 & 75.29 & \textbf{\color[HTML]{0073e6}{76.38}} &\textbf{\color[HTML]{FE0000}{76.52}}\\
		& Stomach              & 67.55 & 69.49 & 54.33 & 70.18 & 70.81 & \textbf{\color[HTML]{0073e6}{71.09}} &\textbf{\color[HTML]{FE0000}{72.26}}\\
		\cmidrule(lr){2-9}
		& \textbf{Average}     & 67.11 & 68.93 & 53.16 & 66.37 & 66.72 & \textbf{\color[HTML]{0073e6}{70.62}}&\textbf{\color[HTML]{FE0000}{71.48}} \\
		\bottomrule
	\end{tabular}
\end{table*}

\section{Experiments}
We trained the proposed model on the MedSeg-HQ dataset, resampling all images to a resolution of 512$\times$512. The dataset was split into training and testing sets with a 9:1 ratio, ensuring patient level independence between the training and testing sets. Model performance is evaluated using the Dice coefficient, covering two types of data: in-domain data (MedSeg-HQ), which measures the model's performance on known data distributions; and out-of-domain data, including TotalSegmentator (MRI) \cite{wasserthal2023totalsegmentator}, SegThor \cite{lambert2020segthor}, and ISLES \cite{hernandez2022isles}, which tests the model's generalization ability. We use the Adam optimizer for model optimization, with an initial learning rate of 1e-4, and apply a learning rate scheduling strategy: the learning rate is reduced by half after the 7th and 12th training epochs. Model training is conducted on 8 NVIDIA RTX 4090 GPUs, with a total batch size of 10. During training, the original SAM architecture remains frozen, and only the newly introduced MoTE and PPT are updated. Unless otherwise specified, all our experiments are conducted on the SAM-Base.

%-------------------------------------------------------------------------
\subsection{Main results}
As shown in Fig.\ref{fig:indomain}, we evaluated the model using two types of interaction modes: point prompts and box prompts. The experimental results demonstrate that although SegMoTE was trained on only about 0.15M mask annotations from the MedSeg-HQ dataset, which is much smaller than the datasets used by other comparative methods, it achieved the best performance across all test sets. This not only validates the effectiveness of our approach but also highlights the importance of the high-quality mask annotations provided by MedSeg-HQ in constructing a robust interactive segmentation benchmark model.

We further conducted zero-shot segmentation experiments on three out-of-domain datasets. The experiments uniformly used boxes as prompts. As shown in Tab.\ref{tab:out}, SegMoTE outperforms the current state-of-the-art models across all metrics. Notably, on the binary classification ISLES dataset, our method improves by 7\% compared to the second best method. On the multi-class SegThor and TotalSegmentator (MRI) datasets, our method surpasses the baseline models by 1\% and 2\%, respectively. It is worth noting that the TotalSegmentator (MRI) dataset includes over 40 categories, and for clarity, we only present the results for abdominal organs.

\subsection{Mask decoder fine-tuning}
To further validate our method's generalization and robustness under full training, we performed end-to-end fine-tuning on SAM's mask decoder. By unfreezing the original decoder parameters and jointly optimizing them with the MoTE and PPT on the MedSeg-HQ dataset, we achieved significant performance improvements. As shown in Tab.\ref{tab:full}, SegMoTE outperforms the baseline by 3\% to 7\% in binary segmentation tasks and demonstrates strong performance in multi-class tasks. These results show that the MoTE and PPT effectively enhance SAM's segmentation capabilities even when the decoder is trained.

\begin{table}[t]
	\centering
	\caption{\textbf{Experimental results on multiple datasets under bounding-box interactions.} Our method involves unfreezing the original decoder parameters and training them along with the MoTE and PPT mechanisms on the MedSeg-HQ dataset.}
	\resizebox{\columnwidth}{!}{ % 宽度设为页面宽度的一半，高度自适应
		\begin{tabular}{lccccc}
			\toprule
			Dataset & SAM & SAM2 & SAM-Med2D & IMIS & Ours \\
			\midrule
			AMOS (CT)            & 76.05 & 78.84 & 80.71 & \textbf{\color[HTML]{0073e6}{83.56}}& \textbf{\color[HTML]{FE0000}{85.16}} \\
			AMOS (MRI)          & 72.63 & 73.46 & 76.55 & \textbf{\color[HTML]{0073e6}{79.72}} & \textbf{\color[HTML]{FE0000}{80.27}}\\
			BTCV               & 77.82 & 78.79 & 80.52 & \textbf{\color[HTML]{0073e6}{82.24}} & \textbf{\color[HTML]{FE0000}{84.51}} \\
			CHAOS (T1)          & 82.67 & 83.30 & 86.14 & \textbf{\color[HTML]{0073e6}{86.92}} & \textbf{\color[HTML]{FE0000}{89.00}} \\
			ISIC2018            & 86.15 & 87.46 & 88.32 & \textbf{\color[HTML]{0073e6}{88.93}} & \textbf{\color[HTML]{FE0000}{93.02}} \\
			SZ-CXR             & 86.72 & 87.58 & 88.72 & \textbf{\color[HTML]{0073e6}{92.03}} & \textbf{\color[HTML]{FE0000}{95.04}} \\
			IELES               & 55.00 & 59.17 & 67.93 & \textbf{\color[HTML]{0073e6}{71.20}} & \textbf{\color[HTML]{FE0000}{78.05}} \\
			SegThor             & 76.55 & 79.81 & 79.06 & \textbf{\color[HTML]{0073e6}{80.96}} & \textbf{\color[HTML]{FE0000}{83.91}} \\
			TotalSegmentator (MRI) & 67.11 & 68.93& 66.72 & \textbf{\color[HTML]{0073e6}{71.01}} & \textbf{\color[HTML]{FE0000}{73.67}} \\
			\bottomrule
		\end{tabular}
	}
	
	\label{tab:full}
\end{table}

%-------------------------------------------------------------------------
\subsection{Ablation Study}

\textbf{Model size comparison.} As shown in Tab.\ref{tab:size}, we compare SegMoTE with SAM, MedSAM, and IMIS in terms of parameter scale. SegMoTE contains only 17M trainable parameters, with the MoTE accounting for 10M and the PPT for 7M, which is only about 1.4\% of the original SAM's total parameter count. Despite the significant reduction in model size, SegMoTE achieves superior segmentation performance while maintaining efficient scalability and strong adaptability. Notably, while MedSAM and IMIS update the entire mask decoder during fine-tuning, SegMoTE outperforms these methods with fewer parameters and significantly lower computational cost, while retaining SAM's general segmentation capability.

\begin{figure}[t]
  \centering
  % 调整图片宽度为 50% 页面宽度
  \includegraphics[width=0.47\textwidth]{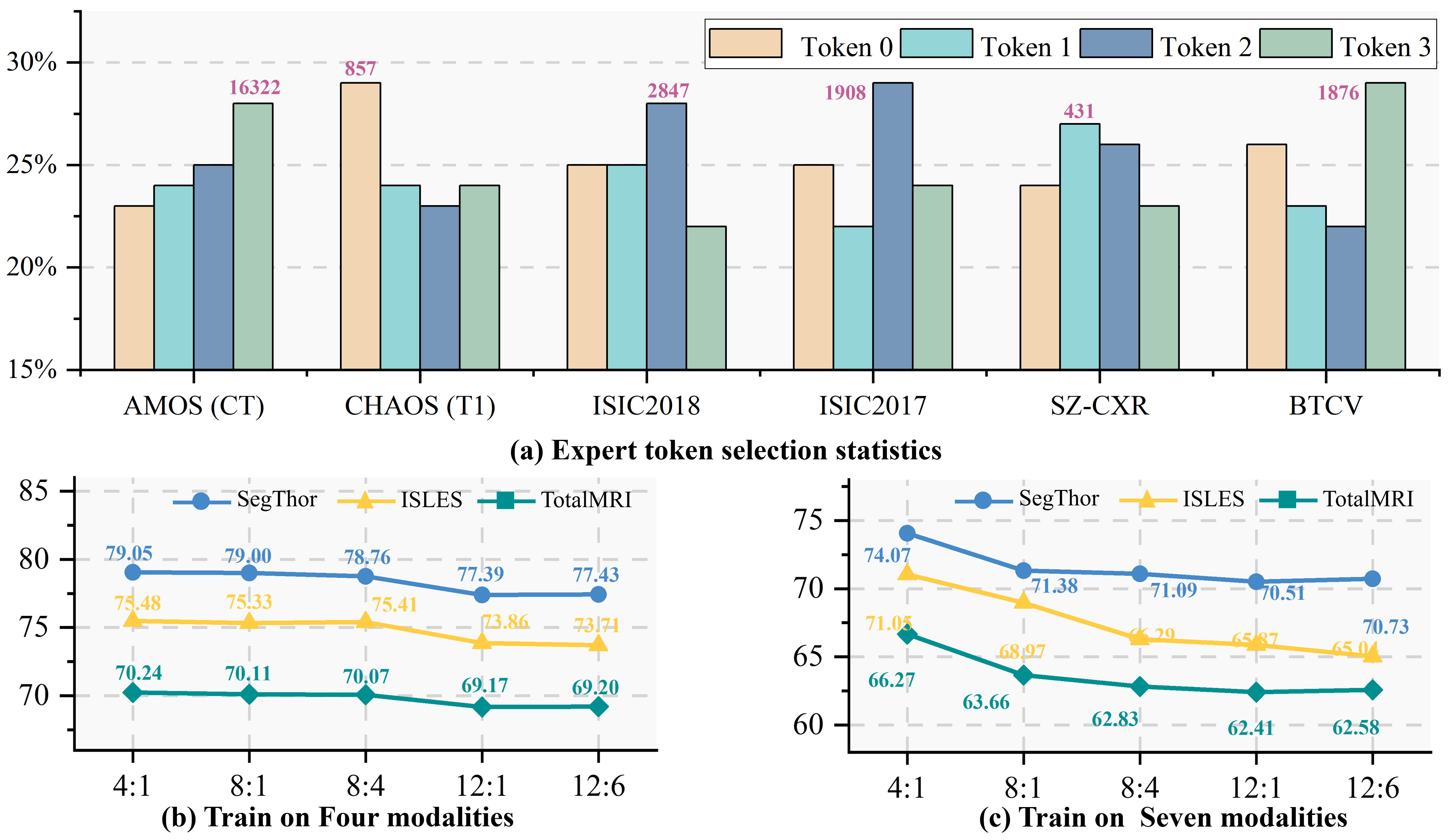}
  \caption{(a) Expert token selection statistics across datasets, the numbers above the bars in the histogram represent the sample size of the current dataset. and (b) (c) experiments with different numbers of experts and activations.}
  \label{fig:select}
\end{figure}

\textbf{Expert Token.} We performed a statistical analysis of the expert token selection mechanism in the MoTE framework. As shown in Fig.\ref{fig:select} (a), under the load balancing constraint, the token selection distribution across datasets is generally balanced, but distinct modality- and task-specific preferences are still observed. Specifically, on the CHAOS (T1) dataset, token 0 is activated more frequently; on ISIC2017/2018, token 2 dominates; in SZ-CXR, token 1 is more active; and on AMOS (CT), token 3 slightly prevails. These consistent selection preferences indicate that different expert tokens have learned discriminative modality-task representations and achieve targeted activation during inference.

We also tested the performance of different expert configurations by training on a subset of MedSeg-HQ and testing on three out-of-domain datasets. Training on four modalities (CT, MRI, MR\_T1W, MR\_T2W), the N:M = 4:1 configuration achieved the best performance, as shown in Fig.\ref{fig:select} (b). When N = 12, performance significantly dropped, as the number of experts exceeded the number of modalities. When training on seven modalities (CT, MRI, MR\_T1W, MR\_T2W, X-ray, Dermoscopy, MR\_FLAIR), N = 4 still outperformed N = 8 and N = 12, as demonstrated in Fig.\ref{fig:select} (c). This shows that four experts are sufficient to capture core features, and additional modalities, such as MR\_FLAIR, can be handled effectively without performance degradation.

\begin{table}[t]
	\centering
	\caption{Comparison of SegMoTE with existing interactive segmentation benchmarks in terms of model size.}
	\resizebox{\columnwidth}{!}{
		\begin{tabular}{lcccc}
			\toprule
			Method & Image Resolution & Learnable Params & Training \#GPU & Batch Size \\
			\midrule
			SAM (Large)            & $1024\times1024$ & 1191 M & A100$\times$128 & 128 \\
			MedSAM (Base)           & $1024\times1024$ & 93 M & A100$\times$20 & 128 \\
			IMIS (Base)           & $1024\times1024$   & 29 M   & RTX4090$\times$72  & 2   \\
			SegMoTE (Base)        & $512\times512$   & 17 M   & RTX4090$\times$8   & 10  \\
			{\hspace*{0.6em}(-MoTE)}  & $512\times512$   & 10 M  & --  & --  \\
			{\hspace*{0.6em}(-PPT)}  & $512\times512$   & 7 M  & --  & --  \\
			\bottomrule
		\end{tabular}
	}
	
	\label{tab:size}
\end{table}

\begin{figure}[t]
  \centering
  % 插入图片，路径注意相对目录是否正确
  \includegraphics[scale=0.2]{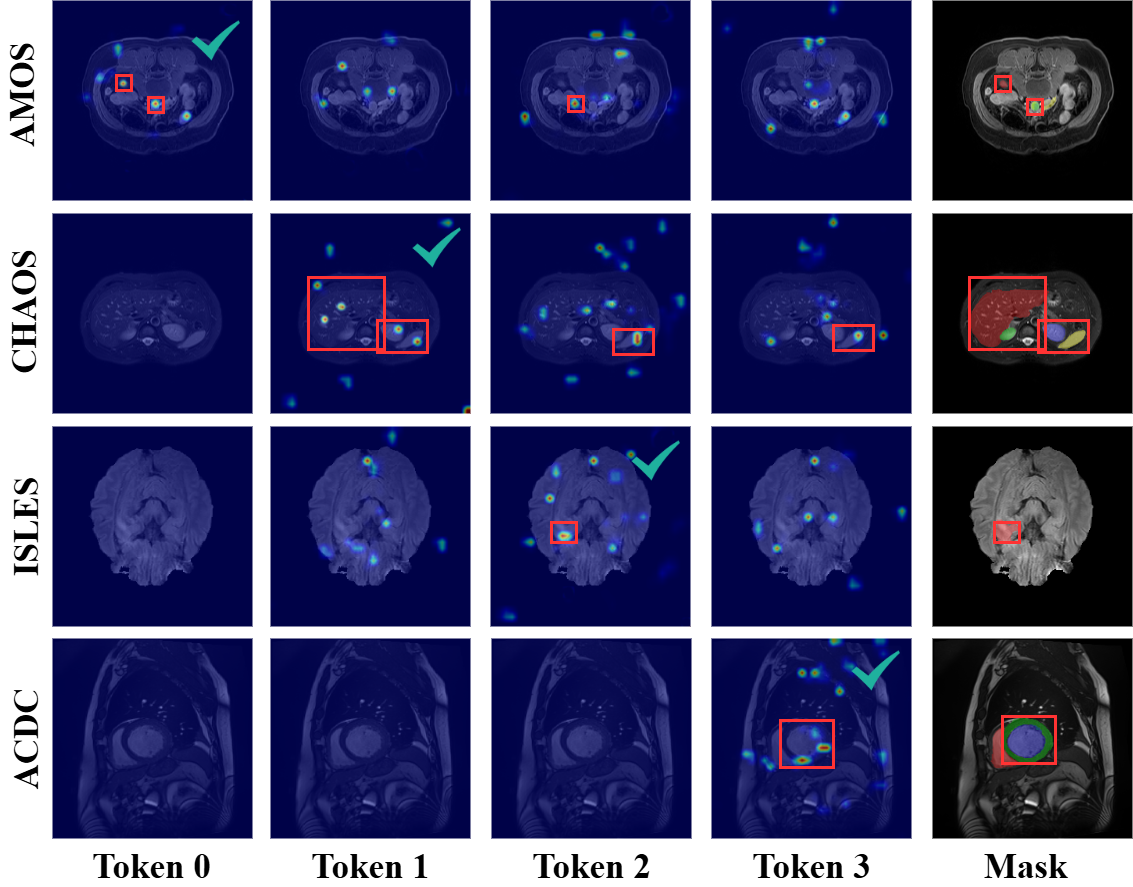}
  \caption{Visualization of expert token routing across datasets in the MoTE framework. Red boxes highlight the target segmentation areas, and the green checkmark indicates the selected token.}
  \label{fig:cam}
\end{figure}

\textbf{MoTE Routing.} To evaluate MoTE’s adaptive token selection across different medical image datasets, we performed a visual analysis of the routing results, as shown in Fig.\ref{fig:cam}. Unlike class activation maps \cite{zhou2016learning,selvaraju2017grad}, our heatmaps show the spatial attention regions corresponding to routed tokens. Since each token’s activation weight is determined by the routing gate, the heatmaps exhibit sparse, discrete activation patterns, which we define as "responsibility regions" to highlight the model’s dynamic attention allocation. Specifically, on AMOS (MRI), token 0 and token 2 both attend to the target region, but token 0 is selected due to its full coverage; on CHAOS (T2), token 1 localizes more accurately while token 0 is inactive; on ISLES, only token 2 covers the target; and on ACDC, token 3 is activated and successfully localizes the target. These results demonstrate MoTE’s ability to adaptively activate expert tokens, showing clear interpretability in path selection.

\begin{table}[t]
	\centering
	\caption{Ablation study on query token number. Analysis of different query token sizes (Q size) under various prompt types. }
	\resizebox{\columnwidth}{!}{
		\begin{tabular}{lcccc}
			\toprule
			Dataset & ISIC2017 & ISIC2018 & SZ-CXR & ISLES \\
			\midrule
			Q=2& \textbf{\color[HTML]{FE0000}{77.62}} & 87.68 & \textbf{\color[HTML]{FE0000}{77.63}} & \textbf{\color[HTML]{FE0000}{65.28}} \\
			Q=4 & 77.23 & \textbf{\color[HTML]{FE0000}{87.79}} & 77.24 & 64.97 \\
			Q=8 & \textbf{\color[HTML]{0073e6}{77.36}} & \textbf{\color[HTML]{0073e6}{87.71}} &\textbf{\color[HTML]{0073e6}{77.51}} & \textbf{\color[HTML]{0073e6}{65.00}} \\
			w/o PPT & 76.66 & 84.87 & 75.20 & 59.00 \\
			\bottomrule
		\end{tabular}
	}
	\label{tab:query}
\end{table}

\textbf{Prompt Token Ablation.} We further evaluated whether automatically generated prompt tokens based on image features can effectively replace manual interactive annotations. As shown in Table \ref{tab:query}, where w/o PPT refers to using points as the prompt method, the model performs excellently on most datasets when the number of learnable queries Q is set to 2 or 8. On in-domain datasets, our method improves performance by 1\% to 3\% compared to traditional interactive methods. On the out-of-domain ISLES dataset, our method surpasses the traditional interactive methods by 6\%, highlighting the significant advantage of using image features for cross-domain generalization. Q = 2 is sufficient for most cases, we used this configuration in our method.
%-------------------------------------------------------------------------

%-------------------------------------------------------------------------

%-------------------------------------------------------------------------

	\section{Conclusion}
We propose SegMoTE, the first model to extend SAM’s “segment anything” capability to general multimodal medical image segmentation while retaining its core functionality, introducing only 17M learnable parameters. SegMoTE integrates a token-level expert routing mechanism with Progressive Prompt Tokenization, enabling adaptive segmentation of multimodal medical images and supporting prompt free inference for binary classification tasks. Trained on the MedSeg-HQ dataset with 0.15M masks, SegMoTE outperforms baseline models trained on large-scale mixed data, both in-domain and out-of-domain. Evaluations across various modalities and anatomical structures show superior generalization performance. This work provides insights for scaling segmentation models with limited annotated data and will explore its effectiveness in 3D data and medical video analysis.

	{
		\small
		\bibliographystyle{ieeenat_fullname}
		\bibliography{main}
	}
	%\input{sec/X_suppl.tex}
	% WARNING: do not forget to delete the supplementary pages from your submission 
	%\input{sec/X_suppl}
	
\end{document}